\documentclass[letterpaper, 10pt, conference]{ieeeconf}      

\IEEEoverridecommandlockouts                              

\usepackage{savesym}
\usepackage{fainekos-macros}
\usepackage{graphics} 
\usepackage{amsmath} 
\usepackage{amssymb}  
\usepackage{xcolor}
\usepackage{graphicx}
\usepackage{multirow}
\usepackage{subcaption}
\usepackage{algorithm}
\usepackage[noend]{algpseudocode}
\usepackage{ifsym}
\usepackage{wasysym}
\usepackage[breaklinks=false, colorlinks=false,linkcolor=black, bookmarks=true, citecolor=black, urlcolor=black,linktoc=all]{hyperref}

\newtheorem{mydef}{Definition}
\newtheorem{myassumption}{Assumption}

\makeatletter
\newcommand{\algmargin}{\the\ALG@thistlm}
\makeatother

\algnewcommand{\parState}[1]{\State%
  \parbox[t]{\dimexpr\linewidth-\algmargin}{\strut #1\strut}}

\savesymbol{Ac}
\usepackage{acronym}
\acrodef{LOS}{line-of-sight}
\acrodef{RRT}{Rapidly-exploring Random Tree}
\acrodefplural{RRT}{Rapidly-exploring Random Trees}
\acrodef{UAV}{Unmanned Aerial Vehicle}
\acrodefplural{UAV}{Unmanned Aerial Vehicles}
\acrodef{AGV}{Autonomous Ground Vehicle}
\acrodefplural{AGV}{Autonomous Ground Vehicles}
\acrodef{CPS}{Cyber-Physical System}
\acrodefplural{CPS}{Cyber-Physical Systems}

\newcommand{\mindist}{\delta_\text{min}}

\title{\LARGE \bf
DEC-LOS-RRT: Decentralized Path Planning for Multi-robot Systems with Line-of-sight Constrained Communication
}

\author{ Victoria Tuck, Yash Vardhan Pant, Sanjit A. Seshia, S. Shankar Sastry
\thanks{$^{1}$Department of Electrical Engineering and Computer Sciences, University of California at Berkeley, CA, USA
	{\tt\footnotesize \{victoria\_tuck, yashpant, sseshia\}@berkeley.edu, sastry@coe.berkeley.edu.}}%
}%

\begin{document}
\maketitle
\thispagestyle{empty}
\pagestyle{empty}

\begin{abstract}

Decentralized planning for multi-agent systems, such as fleets of robots in a search-and-rescue operation, is often constrained by limitations on how agents can communicate with each other. One such limitation is the case when agents can communicate with each other only when they are in \ac{LOS}. Developing decentralized planning methods that guarantee safety is difficult in this case, as agents that are occluded from each other might not be able to communicate until it's too late to avoid a safety violation. In this paper, we develop a decentralized planning method that explicitly avoids situations where lack of visibility of other agents would lead to an unsafe situation. Building on top of an existing \ac{RRT}-based approach, our method guarantees safety at each iteration. Simulation studies show the effectiveness of our method and compare the degradation in performance with respect to a clairvoyant decentralized planning algorithm where agents can communicate despite not being in \ac{LOS} of each other.

\end{abstract}

\section{Introduction}

Autonomous mobile \acp{CPS} like \acp{UAV} and \acp{AGV} are increasingly prevalent and will soon be operating at scale in urban airspaces \cite{nasauam}, roadways, automated warehouses \cite{fork}, etc. One of the challenges in large scale deployment of fleets of such \acp{CPS} is that of scalable and safe planning, especially in the presence of real-world communication constraints. Given both the potentially widespread deployment and the high expected traffic density of these autonomous \acp{CPS} \cite{kockelman2017assessment}, it is unlikely that centralized solutions, like the current air-traffic control system, will work at this scale. Furthermore, the presence of communication constraints also necessitates the development of decentralized approaches for such autonomous systems.

\begin{figure}[tb]
\centering
\includegraphics[width=0.49\textwidth, trim={0cm 0cm 0cm 0cm},clip]{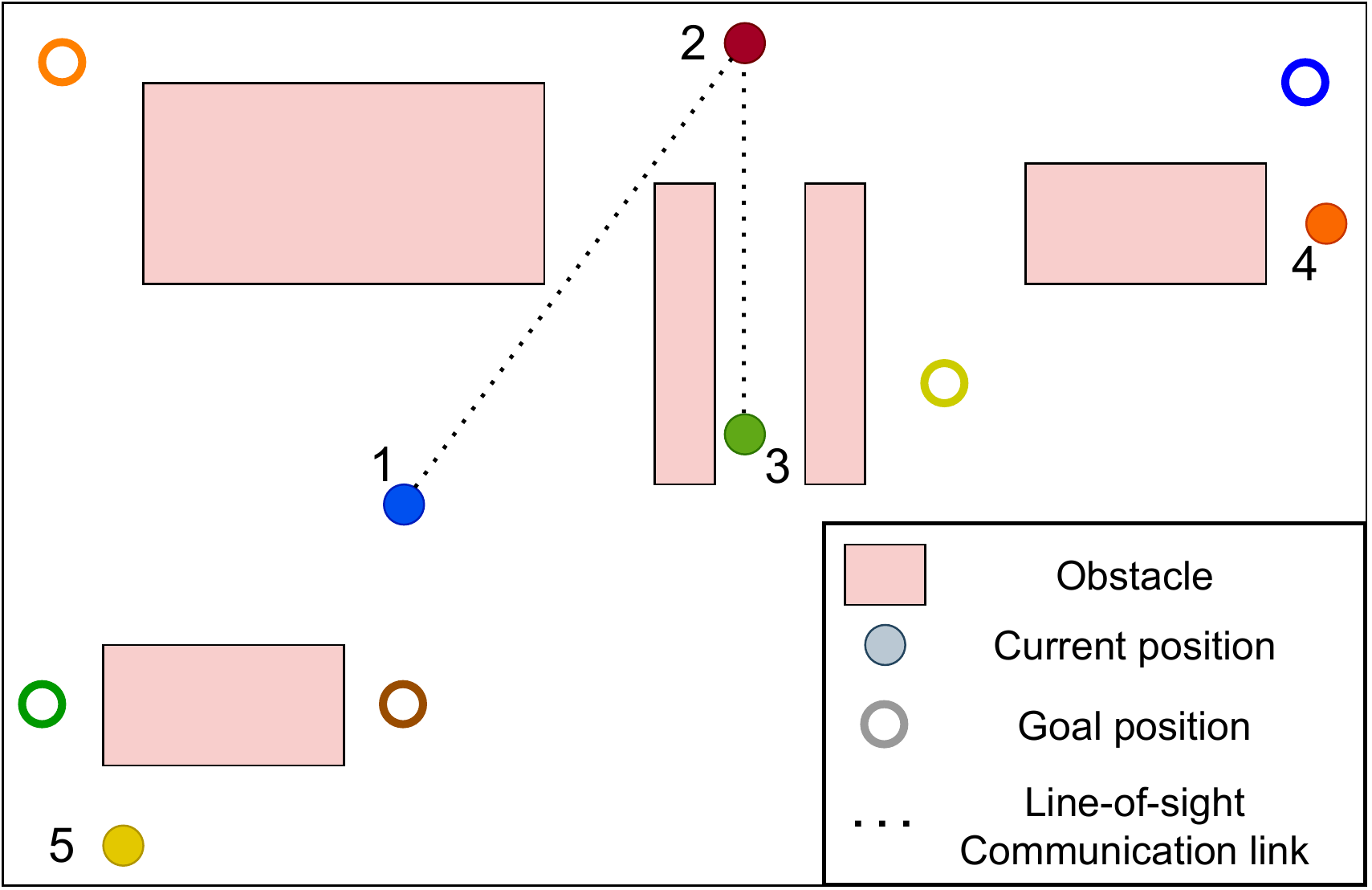}
\vspace{-10pt}
\caption{\small{A high-level visualization of the problem setup. Agents, which can only plan paths for themselves, are tasked with reaching a goal position while avoiding obstacles and other agents. The agents are cooperative and can communicate with other agents that are in line-of-sight (LOS) (e.g., agents 1 and 2, and 2 and 3), and also through multi-hop routing using other nodes in \ac{LOS} as relays (e.g., agents 1, 2 and 3 can communicate with each other). Our approach will allow agents to safely plan movement towards their goal positions with this communication constraint.}}
\label{fig:highlevel}
\vspace{-5pt}
\end{figure}

In this paper, we present a decentralized planning approach for fleets of \acp{CPS} that can only communicate with other agents that are in line-of-sight (\ac{LOS}). This type of communication constraint could arise when the communication link is low-power and cannot be established through solid obstacles---e.g., Bluetooth communication between rescue robots operating in an area with thick walls might be too unreliable for safe and real-time path planning. The need for decentralized planning also arises even when communication constraints are not a major factor e.g., \ac{UAV}-based delivery fleets run by different operators sharing the same airspace or \ac{UAV} operations in urban areas where the airspace density is low. We consider a simplified, lossless communication model where agents in \ac{LOS} can communicate with each other and can create a multi-hop communication network with other agents in \ac{LOS}. With this communication model, we develop a decentralized planning approach for agents tasked with reaching a desired goal position in an \emph{a priori} known workspace with static obstacles. Figure \ref{fig:highlevel} shows an illustration of this setting.


\vspace{3mm}
\noindent \textit{Contributions}
\vspace{3mm}

The main contributions of this work are:

\begin{enumerate}
\item A \textbf{decentralized path planning scheme} for multi-agent systems in an \textit{a priori} known workspace with static obstacles. 
\item The scheme, DEC-LOS-RRT, is designed to be \textbf{aware of the communication constraint} that only agents within \ac{LOS} can communicate with each other. 
\item A proof sketch showing that, by requiring agents to be conservative while avoiding obstacles, the proposed scheme \textbf{guarantees safety}, i.e., no two agents will ever get closer to each other than a pre-defined minimum distance $\mindist$. 
\item A modification of the proposed scheme to reduce the conservatism while still guaranteeing safety. 
\end{enumerate}

Simulation results, including comparisons to the ideal case where all agents can communicate with each other, show the effectiveness of our approach.


\section{Related work}
\label{sec:related}


Decentralized path planning for multi-robot systems has been extensively studied \cite{yan2013survey}, and applied to tasks like information gathering \cite{8260881}, search and rescue \cite{couceiro2017overview}, and shape formation in obstacle-free space \cite{shapeFormation}. 
For general reach-avoid (decentralized) path planning of multiple agents in a workspace with obstacles, \cite{dmaRRT} develops a token passing strategy for coordination that 
ensures safety (collision avoidance) and uses a variant \cite{5175292} of \ac{RRT} \cite{samplingBasedOptimal} for generating the paths. 

 When agents are visible to each other or can communicate, there exist various methods for collision avoidance, e.g., via use of velocity obstacles \cite{orca}, navigation functions \cite{nfmWithGf}, barrier functions \cite{barrierFunc}, or by combining learning-based decision making with decentralized model predictive control \cite{rodionova2020learning}. 
 
 These methods, however, require a connected network to allow broadcasting of information to all agents. Here, we particularly consider the case where only those agents that are in \ac{LOS} can communicate with each other. To deal with such a visibility-based communication constraint, \cite{ensuredNetworkConnectivity, pereira2003decentralized} allow agents to move only in a manner such that the network between subsets of agents remains connected. However, this may not always be possible or optimal given, e.g., goal sets that are spaced out, a large space to move in, or agents that have starting positions outside of line-of-sight from other another. A survey of decentralized path planning with different types of communication constraints between agents can be found in \cite{yan2013survey}. Our method, through being more conservative while avoiding static obstacles, does not require a connected network and yet ensures that agents that are not in \ac{LOS} can avoid imminent collisions.

\section{Problem Statement, definitions and assumptions}
\label{sec:problem}

In this section, we formalize the problem statement that our method solves, state our assumptions, and define relevant terms used in the rest of the paper. 

\subsection{Commonly used notation}
$\mathbb{B}_\delta = \{x \in \mathbb{R}^n. \, ||x||_\infty \leq \delta/2 \}$ represents an inf-norm ball, or square (in planar space) with sides of length $\delta$. We define the $\text{Unique}$ operator on sets as the function that returns the unique sets out of a list of discrete sets; e.g., given sets $A = \{1,2,3\}, \, B = \{3,2,1\}, \,C = \{2,3\}$, $\text{Unique}(A,B,C)$ returns $\{1,2,3\}$ and $\{2,3\}$. $\emptyset$ denotes the empty set, and $|S|$ denotes the cardinality, or number of elements in S. Finally, $\oplus$ and $\setminus$ denote the Minkowski sum and set difference operator, respectively. Note, in the rest of this paper, we use the term \emph{agent} to denote a mobile robot.

\subsection{Agent Models}
Here we define the motion and communication models for our agents. In our setup, there are $N$ agents, indexed by $i,\, i = 1,\dotsc,N$.

\noindent \textbf{Simple robot motion model:} For an agent $i$ , the position state at time $t$ is given by $x_i(t) \in \mathbb{R}^2$. For simplicity, we assume agents have single integrator kinematics of the form $\dot{x}_i(t) = u_i(t)$, where $\dot{x}_i(t) \in \mathbb{R}^2$ is the velocity of the agent at time $t$, and $u_i(t)\in\mathbb{R}^2$ is the control input (or reference velocity) to the agent. Note, while our approach would work with other dynamics as well\footnote{By using other low-level motion primitives in the planning, e.g., paths created by CL-RRT \cite{5175292} instead of RRT as we use here, we can apply this approach to more complex dynamics.}, we consider these simple kinematics to abstract away the low-level motion control which is beyond the scope of this paper. We also make the following simplifying assumption.

\begin{myassumption}
    \label{assumption:dynamics}
    \textit{Instantaneous braking:} Agents can instantaneously brake, i.e., can set $\dot{x}_i(t) = 0$ at any $t$. 
\end{myassumption}


\noindent \textbf{Communication model:} We study the problem of communication constrained decentralized path planning for multi-robot systems in an \emph{a priori} known workspace. Agents can directly communicate with other agents that are visible, i.e., not occluded by obstacles (see Figure \ref{fig:highlevel}).

\begin{mydef}
\label{def:workspace}
\textit{Workspace and obstacles}: The workspace $W \subset \mathbb{R}^2$ is a compact set. Static obstacles $O_m \subset W,\, \forall m = 1,\dotsc L$ are axis-aligned rectangles. All agents have \emph{a priori} knowledge of the sets $W$ and $O_m$.
\end{mydef}

\begin{myassumption}
    \label{def:line-of-sight}
    \textit{Line-of-Sight (LOS) communication:} Two agents with positions $x_i$ and $x_j$ can directly communicate with each other if $x_i$ is within \ac{LOS} of another agent at position $x_j$, i.e., $x_i (1 - \lambda) + x_j\lambda \notin  \bigcup_{m}{O_m}  \, \forall \lambda \in [0,1],\, \forall m$. That is, a point on a line segment between the positions of the agents cannot also be a point inside an obstacle. 
\end{myassumption}

\begin{mydef}
\label{def:neighbors}
\textit{Visible neighbors:} Agents within LOS of an agent $i$ are called \emph{visible neighbors}, or \emph{neighbors} of agent $i$. The \emph{neighborhood} of an agent $i$ is the set of agents $N(i) = \{j | \, x_i (1 - \lambda) + x_j\lambda \notin \bigcup_{m}{O_m} \} \,\forall \lambda \in [0,1],\, \forall m $.
\end{mydef}

\begin{mydef}
    \label{def:visible-subgraph}
    \textit{Agent communication subgraphs:} An agent $i$'s communication subgraph $S$, includes all neighbors and agents in the subgraphs of neighbors. $S(i) = \{i \cup N(i) \cup \bigcup_{j \in N(i)}{S(j)}\}$.
\end{mydef}

\begin{mydef}
    \label{def:subgraph-set}
    \textit{Communication subgraphs:} The unique agent communication subgraphs represent what we call \emph{communication subgraphs}, or just \emph{subgraphs}, given by  $\mathcal{S} = \{S_1,\dotsc,S_P\} = \text{Unique}(\{S(i) \}_{i=1}^N)$. These represent the sets of agents connected with each either directly via \ac{LOS} (neighbors) or through neighbors of neighbors. 
\end{mydef}

\begin{exmp}
The set of all communication subgraphs for the instance shown in Figure \ref{fig:highlevel} is $\mathcal{S} = \{\{4\},\,\{5\}, \,\{1,2,3\}\}$.
\end{exmp}

Note, $\bigcup_{p=1}^PS_p = \{1,\dotsc,N\}$ and $S_p \bigcap S_{p'}= \emptyset \,\forall p' \neq p$. Finally, we make the following assumption that gives us the full communication model considered here:

\begin{myassumption}
    \label{assumption:communication}
    Agents can communicate with other agents in their communication subgraphs, and the communication between agents is lossless and has no latency.    
\end{myassumption}

The above assumption implies that agents can communicate with other agents that are in \ac{LOS}, i.e., neighbors, and also with neighbors of neighbors via a multi-hop communication network.

\subsection{Problem Statement}
\label{sec:problemstatement}

For a system with $i=1,\dotsc,N$ agents with the agent and communication models described above, we aim to develop a decentralized planning approach where each agent (within a bounded time interval $[0,T]$) wants to satisfy the following requirements:

\begin{enumerate}

\item \emph{Reach goal state:} Given a desired final position for each agent $x_{if}$,  
\begin{equation}
\label{eq:goalstate}
\exists t_{if} \in [0,T]\, \text{s.t.}\, x_i(t_{if}) \in x_{if}\oplus\mathbb{B}_\epsilon\end{equation} 

where $\epsilon$ is a desired tolerance. 
\item \emph{Avoid static obstacles:}
\begin{equation}
\label{eq:avoidobstacle}
 x_i(t) \notin \bigcup_m O_m \forall t\in[0,T], \forall i=1,\dotsc,N
 \end{equation}
 
\item \emph{Avoid other agents:} Agents must not get closer than a minimum distance $\mindist$ to each other:
\begin{equation}
\label{eq:minseparation}
||x_i(t)-x_j(t)||_{\infty} \geq \mindist \forall t \in [0,T], \, \forall i \neq j
\end{equation}

\end{enumerate}

Here, \eqref{eq:avoidobstacle} and  \eqref{eq:minseparation} are the \emph{safety constraints} that all agents should always satisfy. Next, we define the following more conservative version of an obstacle that all agents must avoid for purposes of the algorithm explained in section \ref{sec:solution}:



\begin{mydef}
    \label{def:deltaobs}
    \textit{$\delta$-Obstacle:} For a given rectangular obstacle $O_m$, a $\delta$-Obstacle is defined as $O_{\delta,m} = O_m \oplus \mathbb{B}_\delta$. This corresponds to virtually \emph{expanding} the obstacle by an inf-norm ball of side $\delta$. See figure \ref{fig:adaptivedelta} for a visualization.
\end{mydef}

Finally, we make the following assumption on the initial positions of all agents:

\begin{myassumption}
\label{assumption:safeints} \textit{Safe initialization:} At time $t=0$, all agents have positions $x_i(0) = x_{i0} \notin \bigcup_m O_{\delta,m}$, i.e., outside of all $\delta$-obstacles and also satisfy \eqref{eq:minseparation}.
\end{myassumption}


\section{DEC-LOS-RRT for decentralized path planning with LOS communication}

\label{sec:solution}
    
This section explains our approach and gives a proof sketch of its safety guarantee. 
\subsection{Overview of Approach}

Our approach involves the following steps: 

\begin{enumerate}
\item \textbf{Decentralized path planning for agents in \ac{LOS}:} Agents plan paths such that all agents in the same visibility subgraph can safely navigate towards their goal while avoiding each other and the $\delta$-obstacles. We use DMA-RRT, an off-the-shelf RRT-based algorithm \cite{dmaRRT}, for decentralized path planning for each agent in a visibility subgraph. 

\item \textbf{Avoiding blind-spots in the workspace and safety between agents not in \ac{LOS}:} We do this by planning paths for each agent such that they are no closer than $\delta$ distance (in the inf-norm sense) to obstacles in the workspace (see definition \ref{def:deltaobs}). As will be shown in lemma \ref{thm:lemma1}, we find a value for $\delta$ as a function of $\mindist$  such that that it is ensured, by construction, that agents from different communication subgraphs (and hence not in \ac{LOS}) never enter a situation where they are closer than the minimum desired separation $\mindist$. 

\item \textbf{Safety between new agents emerging in \ac{LOS}:} Finally, when agents not previously in the same visibility subgraph are now in \ac{LOS}, these agents a) perform an emergency braking maneuver if needed, and b) recompute visible subgraphs. Agents in each visibility subgraph go back to using DMA-RRT for decentralized path planning.
\end{enumerate}

As will be formalized later in this section in theorem \ref{thm:mainresult}, iterative application of these steps ensures that recursively, all agents are safe (avoid obstacles and each other) at each time step. We now present the main algorithm that we develop to solve the problem stated in section \ref{sec:problemstatement}.

\subsection{The DEC-LOS-RRT Algorithm}
This section includes a more detailed description of the algorithm and the theory behind it. The main algorithm is shown in Algorithm \ref{alg:Adjusted_dma_rrt} and is able to guarantee safety at each time step, as is proved later in this section. This version of the algorithm uses Decentralized Multi-Agent Rapidly-exploring Random Tree (DMA-RRT) \cite{dmaRRT}\footnote{Another decentralized multi-agent path planning algorithm could be used in place of DMA-RRT if it provides the same guarantees.} as a building block for planning for the agents that are in the same communication subgraph (see Definition \ref{def:visible-subgraph}). 
Some commonly used notation is $x$ for an agent's position, $k$ for the algorithmic step, $t$ for time and $dt$ for the time step. $x$ may be indexed at a time $t$ with $x(t)$ or at a time equal to the algorithmic step $k$ with $x[k]$. A bolded $\textbf{x}$ is a vector of the combined states of all N agents when without subscript and of all $|S|$ agents in a subgraph $S$ with subscript $S$ such as $\textbf{x}_S$. We assume the outer loop of the algorithm (increments of $k$) occurs once per second and that $dt$ is small enough such that emergency stops happen as soon as agents are in LOS. The set of waypoints for an agent $i$ is the set $w_{0:M,i} = \{x_i[k], \, k = 0,\dotsc,M\}$, which is bolded when it includes more than one agent such as $\textbf{w}_{0:M,S}$ for all agents in the subgraph $S$. 

\begin{algorithm} [!htb]
\caption{DEC-LOS-RRT}\label{new_dmarrt}
    \begin{algorithmic} [1]
        \State $t = 0, k = 0, w_{0:M,i} = \{x_{i0}\} \forall i=1,...N$
        \State $\mathcal{S} = \text{findSubgraphs}(\textbf{x}[k])$
        \For {$S \in \mathcal{S}$}
            \State $\{bids_{k,S}\} = DMA\_RRT\_INIT(S)$
        \EndFor
        \While{$\exists \ i \in \{1,...N\}$ st $x_i[k] \notin x_{if} \oplus\mathbb{B}_\epsilon$ and $k < M$}
            \State $k \gets k + 1$
            \For{$S \in \mathcal{S}$}
                \parState {%
                    $\{winner_{k,S}, \textbf{w}_{k:M,S}, bids_{k,S}\} = DMA\_RRT\_STEP(S, bids_{k-1,S})$ }
            \EndFor
            \While {$ t < k $}
                \State $ t \gets t + dt$
                \State Track $w_{k,i}$ in $\{ \mathbf{w}_{k:M,S}$ for all i in all S in $\mathcal{S}\}$
                \State $\mathcal{S}_{new} = \text{findSubgraphs}(\textbf{x}(t))$ 
                \If {$\mathcal{S}_{new} \neq \mathcal{S}$}
                    \For {$S \in \mathcal{S}_{new} \setminus \mathcal{S}$}
                        \State $\dot{x_i} = 0 \quad \forall i \in S$
                        \State $\textbf{w}_{k:M,S} = \{\textbf{x}_S(t)$\}
                        \State $\{bids_{k,S}\} = DMA\_RRT\_INIT(S)$
                        \State $S \gets S_{new}$
                    \EndFor
                \EndIf
            \EndWhile
        \EndWhile
    \end{algorithmic}
    \label{alg:Adjusted_dma_rrt}
\end{algorithm}


\textbf{DMA-RRT \cite{dmaRRT}-based planning for agents in a visibility subgraph:}
The DMA-RRT algorithm \cite{dmaRRT}, which we use for decentralized planning for agents in a visibility subgraph, ensures that all agents in the visibility subgraph are no closer than $\mindist$ to each other and $\delta$ to the static obstacles. DMA-RRT is a token-passing algorithm where agents place bids to receive the passed token. The initial token holder is chosen randomly. Each agent continuously searches for a better (with respect to a cost function like path length or travel time) set of waypoints using RRT. The bid they pass is based on the Potential Path Improvement (PPI) that the agent could gain by using the updated set of waypoints as opposed to their current ones. The agent currently holding the token decides the winner (agent with the highest bid) of each round, announces that to all agents, and passes the token to the winner. Once an agent has received the token, it is allowed to assign its actual waypoints to this desired set of waypoints at which point all other agents must take this path into consideration as a constraint on their path planning, and the process is repeated.

This algorithm assumes a fully connected, lossless network with negligible delays, which is consistent with Assumption \ref{assumption:communication}. Note that while agents in a communication subgraph may not have a LOS between every agent, the subgraph will be fully connected none-the-less because of the allowed communication hops. We also make an assumption about the initial set of waypoints of the agents.

\begin{myassumption}
    \label{assumption:waypoints}
    Agents begin with a set of waypoints $\mathbf{w}_{0:M}$ that fulfills all safety constraints.
\end{myassumption}

The safety constraints require all agents to avoid other agents in their subgraph with a distance $\delta_\text{min}$ and $\delta$-obstacles as defined in Section \ref{sec:problem}. 
Therefore, a valid initial set of waypoints (sequence of $M+1$ waypoints for each agent) that fulfills this assumption might simply correspond to the agents being stopped at their initial position $x_{i0}$ (also see Assumption \ref{assumption:safeints}) as is used in line 1 of Algorithm \ref{alg:Adjusted_dma_rrt}.

%
The following are the guarantees that DMA-RRT provides, applicable for agents in a subgraph, as long as no agent from another subgraph appears in \ac{LOS}.
\begin{theorem}\cite{dmaRRT}
\label{thm:dmarrt}
Given a set of $N'$ cooperative agents that can communicate with each other at all times and a corresponding set of inter-agent safety constraints satisfying all assumptions, if the initial set of waypoints $\{w_{0:M, i} | \forall i, i = 1,...,N' \}$ satisfies all safety constraints, all future sets of waypoints $\{w_{k:M, i} | \forall i, i = 1,...,N'\} \forall k = 0,...,M $ and the associated trajectories will satisfy all safety constraints.
\end{theorem}

Here, $w_{0:M, i}$ is the sequence of $M+1$ position waypoints for agent $i$. Each subgraph of agents runs a separate instance of DMA-RRT while continuously checking to see if the subgraph has changed i.e., a new agent joins or leaves the subgraph, e.g., by appearing in \ac{LOS} when it previously was not. If it does, agents in this subgraph stop, update their future waypoints, and start a new instance of DMA-RRT.

\textbf{Use of $\delta$-Obstacles:} In DEC-LOS-RRT, we require that agents avoid $\delta$-obstacles, which are defined in definition \ref{def:deltaobs}. The added conservatism allows us to guarantee safety of all agents with respect to both static obstacles and other agents, including those that are not in the same communication subgraph. This is formalized in the following lemma:

\begin{lemma}
\label{thm:lemma1}\textit{$\delta$-Obstacles and avoiding collisions with agents not in \ac{LOS}:}
For any two agents $i$, $j$ such that they: 1) $x_j \notin S(i)$, i.e., are not in the same communication subgraph, and 2) $x_i(t), \, x_j(t)\ \in W \setminus \bigcup_m O_{\delta,m}$, i.e., are not in the $\delta$-obstacles, then, $||x_i(t)-x_j(t)||_\infty > \delta_\text{min}$ if $\delta>\frac{1}{2}\delta_\text{min}$.
\end{lemma}

This lemma states that two agents not in \ac{LOS} and not in the same subgraph will not be closer than the minimum distance $\mindist$ to each other as long $\delta > (1/2)\mindist$. The proof sketch for this is in the appendix and follows from a simple geometric construction of the $\delta$-obstacle and agent positions. 

\textbf{Interrupting and restarting DMA-RRT.}
In algorithm \ref{alg:Adjusted_dma_rrt}, DMA-RRT is initialized for each subgraph [line 4] and then reinitialized [line 17] if a subgraph calculation [line 12] and corresponding check [lines 13/14] finds that the subgraph an agent belongs to has changed. At every iteration $k$, DMA-RRT will run one step [line 8]. Due to assumption \ref{assumption:waypoints}, each time a subgraph changes [line 14], it is necessary for waypoints to be reinitialized [line 16] because the set of waypoints each agent in a subgraph has before the subgraph changes may not still be valid with the addition of an agent. However, due to lemma \ref{thm:lemma1}, the agents' current positions will be safe. Therefore, each time DMA-RRT is reinitialized [line 17] of algorithm \ref{alg:Adjusted_dma_rrt}, the initial set of waypoints is reset to be a stopped state at each agent's current position to fulfill assumption \ref{assumption:waypoints} [lines 15/16].

 
Finally, given Lemma \ref{thm:lemma1}, the guarantees from DMA-RRT (Theorem \ref{thm:dmarrt}), the ability for agents to continuously check and update subgraphs, and the ability of the agents to perform emergency braking as in assumption \ref{assumption:dynamics}, algorithm \ref{alg:Adjusted_dma_rrt} guarantees safety at each time step in a recursive manner:


\begin{theorem}[Recursive safety]
\label{thm:mainresult}
Given $N$ cooperative agents, and the set of all unique visibility subgraphs $\mathcal{S}$ where each agent $i \in \{1,\dotsc,N\}$ starts with a valid initial set of waypoints $w_{0:M,i}$ that satisfy the safety constraints with respect to the $\delta$-obstacles and other agents in its subgraph $S$, then using Algorithm \ref{alg:Adjusted_dma_rrt} with agents avoiding $\delta$-obstacles, all future positions for all agents $x_i(t) \forall t\in[0,T],\, \forall i=1,...,N$ will satisfy the safety constraints \eqref{eq:avoidobstacle} and \eqref{eq:minseparation}.
\end{theorem}

\textit{Proof sketch:} For each subgraph, agents will use DMA-RRT until the nodes of the subgraph change. The main DMA-RRT theorem states that starting from an initial set of valid paths and following the algorithm will lead to valid future paths. Assume an initial valid path is defined by the waypoints $\mathbf{w}_{0:M,S}$ for each subgraph $S$. If at a point in time $t$, an agent in subgraph A sees an agent in subgraph B and immediately stops, from lemma \ref{thm:lemma1}, the agents will be at least $ \mindist$ away from each other and therefore still fulfill inter-agent constraints. The agents will also fulfill any other constraints due to the safety guarantees of DMA-RRT (theorem \ref{thm:dmarrt}). Therefore, the positions $x_i(t) \, \forall i \in S_{new}$, where $S_{new}$ is the subgraph including the new agent and others it is connected to, will still fulfill all constraints and can be used as a stopped point for a new instantiation of DMA-RRT for the new subgraph. This can be applied to all subgraphs separately and can be applied inductively to show all future positions will satisfy all safety constraints.

\textit{Remark:} Similar to DMA-RRT, our algorithm also guarantees safety (\eqref{eq:avoidobstacle} and \eqref{eq:minseparation}) and only makes a best effort to reach the goal state \eqref{eq:goalstate} within the given upper bound on time $T$. In practice, since we use RRT as the lower-level planner, our approach does manage to find paths that reach the goal state for almost every agent as seen in section \ref{sec:simulations}.

\begin{figure}[tb]
\centering
\includegraphics[width=0.49\textwidth, trim={0cm 0cm 0cm 0cm},clip]{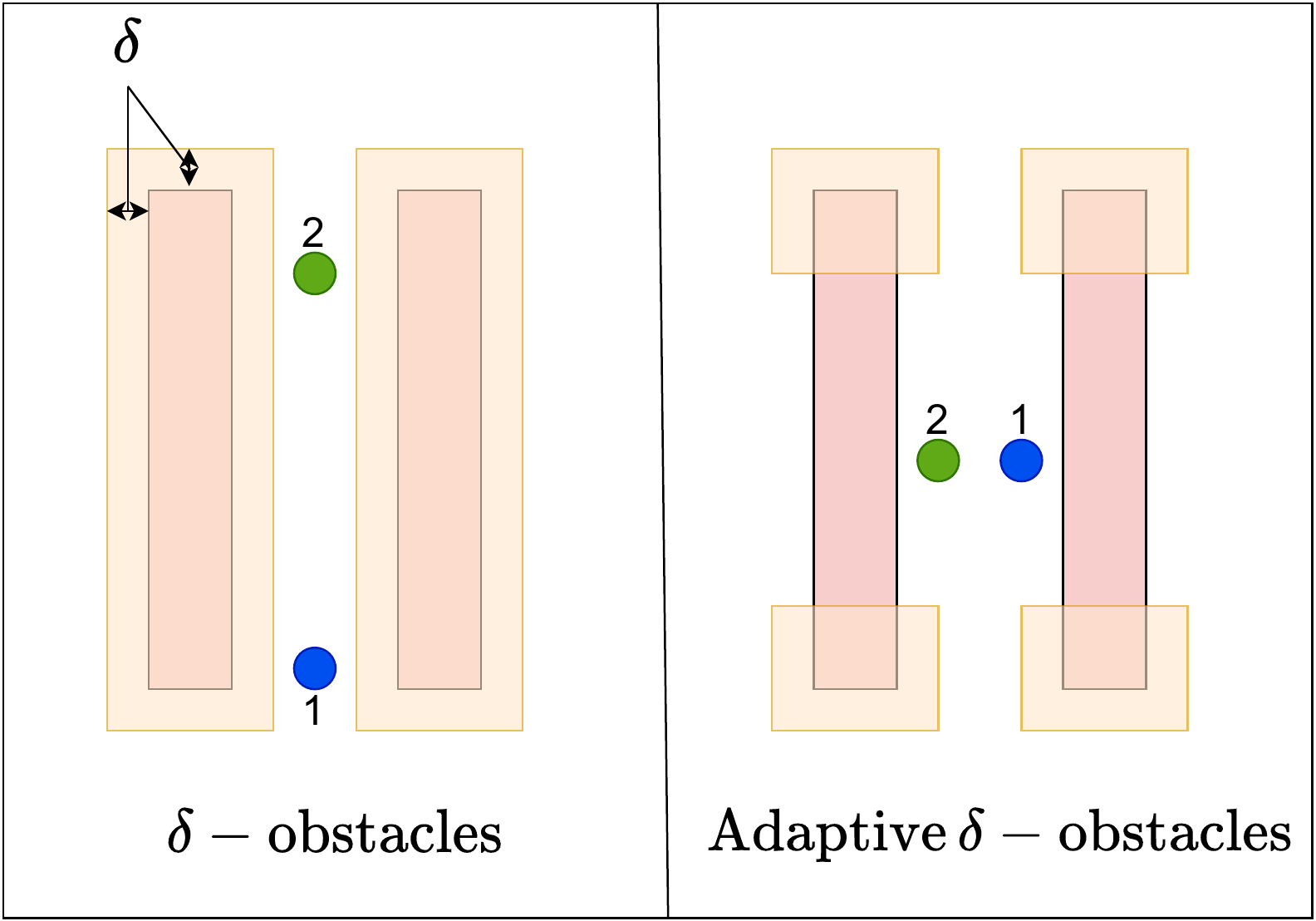}
\vspace{-10pt}
\caption{\small{Left: $\delta$-obstacles obtained by virtually expanding the obstacles as in definition \ref{def:deltaobs}. Right: Adaptive $\delta$-obstacles, where not all of the obstacle is expanded by $\delta$. Note, this expansion is only for the purpose of the planner, and the \ac{LOS} is determined by the occlusions caused by the physical obstacles. }}
\label{fig:adaptivedelta}
\vspace{-5pt}
\end{figure}

\subsection{Adaptive obstacle expansion to reduce conservatism}
\label{sec:lessconservative}

In some situations, virtually expanding the entire obstacle by $\delta$ might be unnecessary, e.g., consider the case shown in Figure \ref{fig:adaptivedelta}. The $\delta$-obstacles would result in an overly narrow corridor where the two agents $1$ and $2$ could not pass through side-by-side. On the other hand, virtually expanding only some parts of the obstacle by $\delta$ would allow the agents to pass side-by-side, while also ensuring that lemma \ref{thm:lemma1} still holds. This adaptive expansion of an obstacle would result in a union of rectangular obstacles as shown in Figure \ref{fig:adaptivedelta}. 
We can use these adaptive $\delta$-obstacles as the obstacles that are to be avoided and still retain the safety guarantees while also reducing the conservatism in our approach.

\section{Simulation Results}
\label{sec:simulations}

In this section, we evaluate our method on a given workspace with a varying number $N$ of agents. We also compare our method to the best-case performance of using purely DMA-RRT, which assumes that all $N$ agents can talk to each other, unlike our approach which respects the communication constraints in assumption \ref{assumption:communication}. For the comparison, both the DEC-LOS-RRT algorithm and DMA-RRT algorithm are run with the same starting and goal positions. Note, due to the lack of communication constraints for the DMA-RRT baseline it provides a best-case scenario that we use to evaluate the degradation in DEC-LOS-RRT performance (due to constrained communication).

\subsection{Simulation setup}
\noindent\textbf{Implementation details:} The simulations were run on a computer running macOS Mojave with a 2.6 GHz Intel Core i7 and 16 GB RAM. The implementation for our method, DEC-LOS-RRT, and for DMA-RRT were done in Python, building on top of a RRT* implementation from PythonRobotics \cite{pythonrobotics}. For the decentralized planning of agents in a communication subgraph, a token passing strategy based on the agent that is next in order was used as a simplification instead of comparing PPI. Additionally, a new (RRT*) tree is started for each agent at each DMA-RRT step and is run for up to 150 iterations.

\begin{figure}[tb]
\centering
\includegraphics[width=0.49\textwidth, trim={0cm 0cm 0cm 0cm},clip]{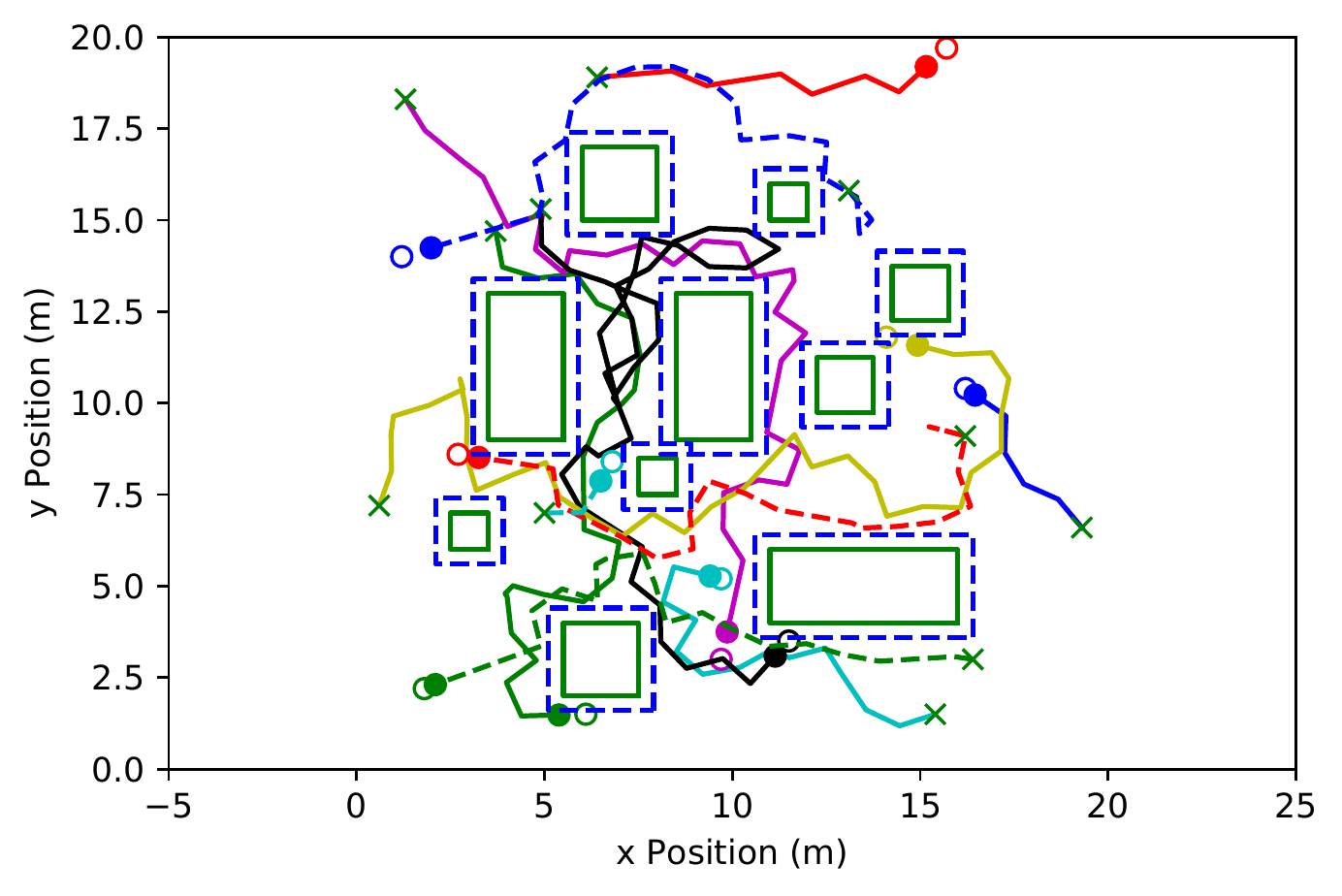}
\vspace{-10pt}
\caption{\small{11 agents traveling to their goal positions while running DEC-LOS-RRT. All agents satisfy the safety constraints. The boxes with solid lines represent the obstacles, and the dashed lines around them represent the $\delta$-obstacles. \Circle~represents the agent goal states, \CIRCLE~the agent positions at the last time step (coinciding with \Circle~after reaching the goal state), and $\times$ the initial positions. A video of this simulation can be found at \protect{\url{https://www.youtube.com/playlist?list=PLT1tyNQyXbkKDY7uH1XY7d7O3IsM_dlCu}}}}
\label{fig:workspace1}
\vspace{-5pt}
\end{figure}

\noindent\textbf{Workspace and simulation parameters:} The workspace with dimensions $[0, 20]$ in meters along both axes can be seen in Fig. \ref{fig:workspace1}. It has three rectangular obstacles and seven square obstacles.  Agents have starting and goal positions chosen randomly and located outside of the $\delta$-obstacles. $\epsilon$ for the goal set is set to be 1 m. We run simulations for up to $N=11$ agents with randomly chosen start positions. In Fig. \ref{fig:workspace1} agents have start positions of (agent\#-[x,y]): 1-[19.3, 6.6], 2-[3.7, 14.7], 3-[6.4, 18.9], 4-[15.4, 1.5], 5-[1.3, 18.3], 6-[4.9, 15.3], 7-[0.6, 7.2], 8-[13.1, 15.8], 9-[16.4, 3.0], 10-[16.2, 9.1], 11-[5.0, 7.0] with goal positions of: 1-[16.2, 10.4], 2-[6.1, 1.5], 3-[15.7, 19.7],  4-[9.7, 5.2], 5-[9.7, 3.0], 6-[11.5, 3.5], 7-[14.1, 11.8], 8-[1.2, 14.0], 9-[1.8, 2.2], 10-[2.7, 8.6], 11-[6.8, 8.4]. Agents must avoid each other with at least of $\mindist = $ 0.6m, and, in accordance with lemma \ref{thm:lemma1}, the smallest desired $\delta$ is 0.3 + $\epsilon_c$m  where $\epsilon_c$ is a small positive constant. However, in practice, because $dt$ cannot fulfill the above assumption that it is infinitesimally small, we set $\delta$ = 0.4 + $\epsilon_c$m to account for the distance an agent can travel in $0.1$s in our simulation. The DEC-LOS-RRT iteration $k$ is incremented every $1$s, and $dt$ is set to $0.1$s, so the agents move and check subgraphs at a frequency of 10 Hz.

\begin{figure}[tb]
\centering
\includegraphics[width=0.49\textwidth, trim={0cm 0cm 0cm 0cm},clip]{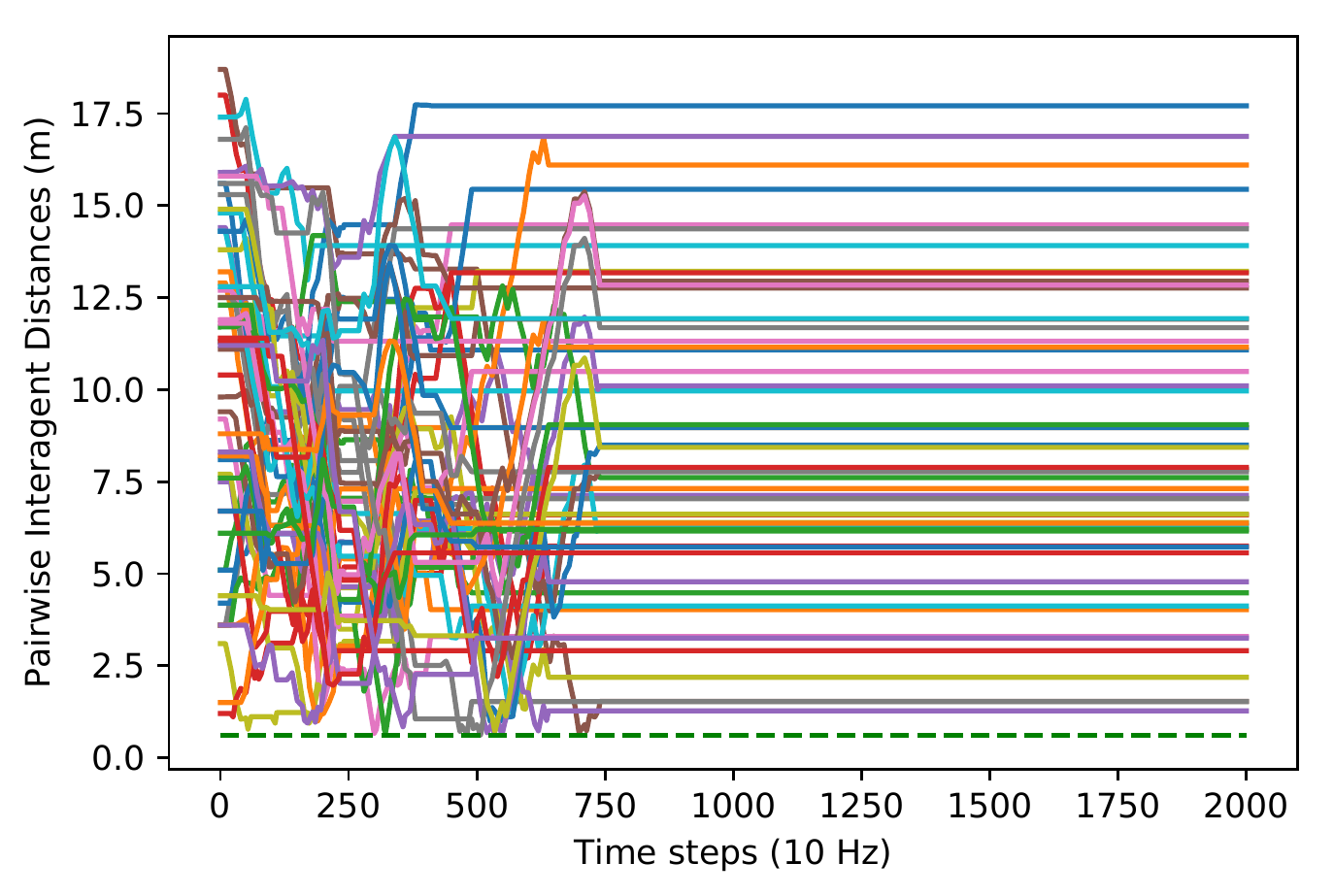}
\vspace{-10pt}
\caption{\small{Inter agent distances with no violations for 11 agents. The dashed line shows $\mindist$, which the agents do not cross.}}
\label{fig:mindists}
\vspace{-5pt}
\end{figure}


\subsection{Main Results}
\begin{figure}[tb]
\centering
\includegraphics[width=0.49\textwidth, trim={0cm 0cm 0cm 0cm},clip]{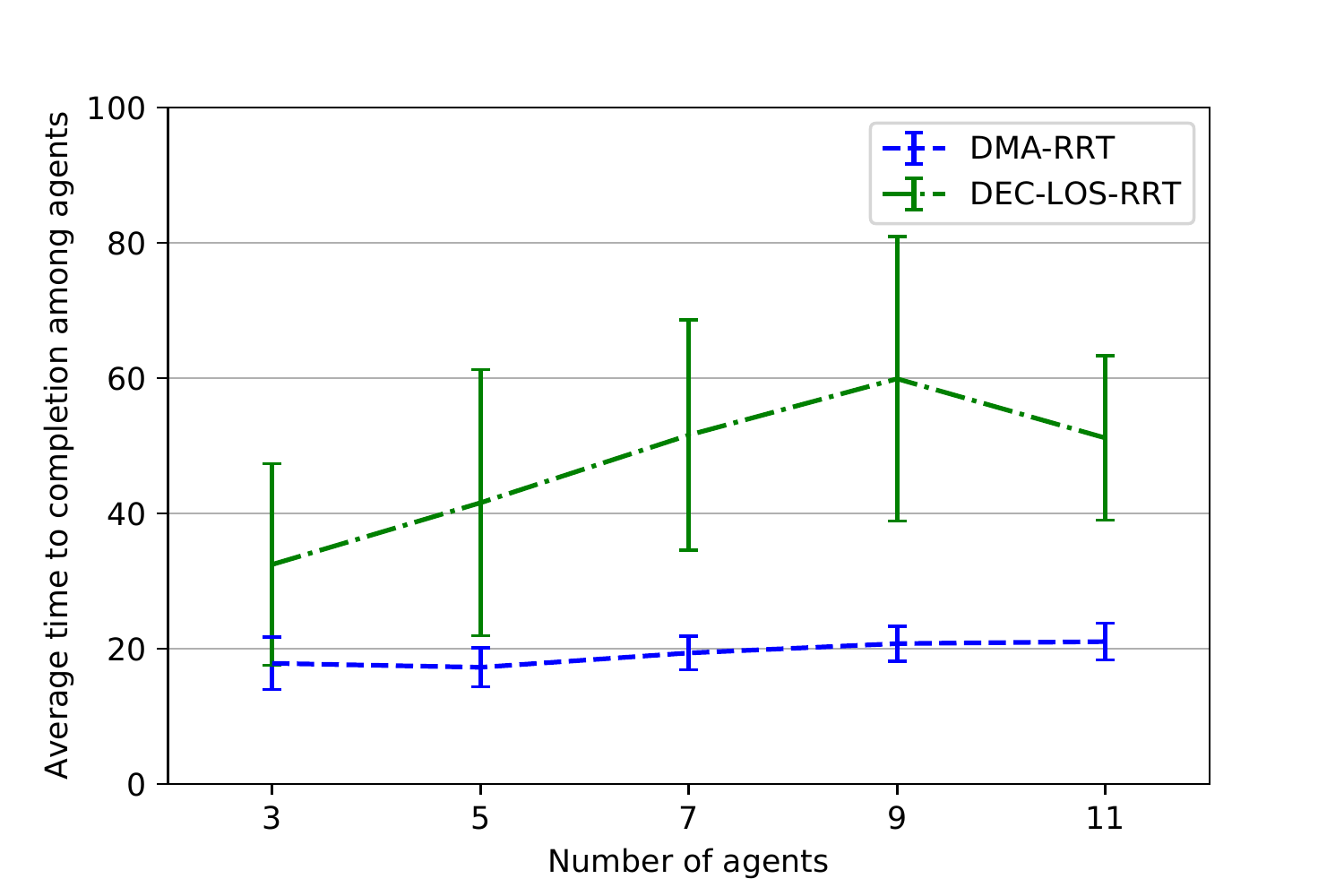}
\vspace{-10pt}
\caption{\small{Average time for an agent to reach their goal vs the number of agents running the algorithm total. DEC-LOS-RRT takes longer for agents to reach their goal since the constrained communication model requires emergency braking and re-computation of agent paths when communication subgraphs change. One standard deviation is shown.}}
\label{fig:comparison_time}
\vspace{-5pt}
\end{figure}

\begin{figure}[tb]
\centering
\includegraphics[width=0.49\textwidth, trim={0cm 0cm 0cm 0cm},clip]{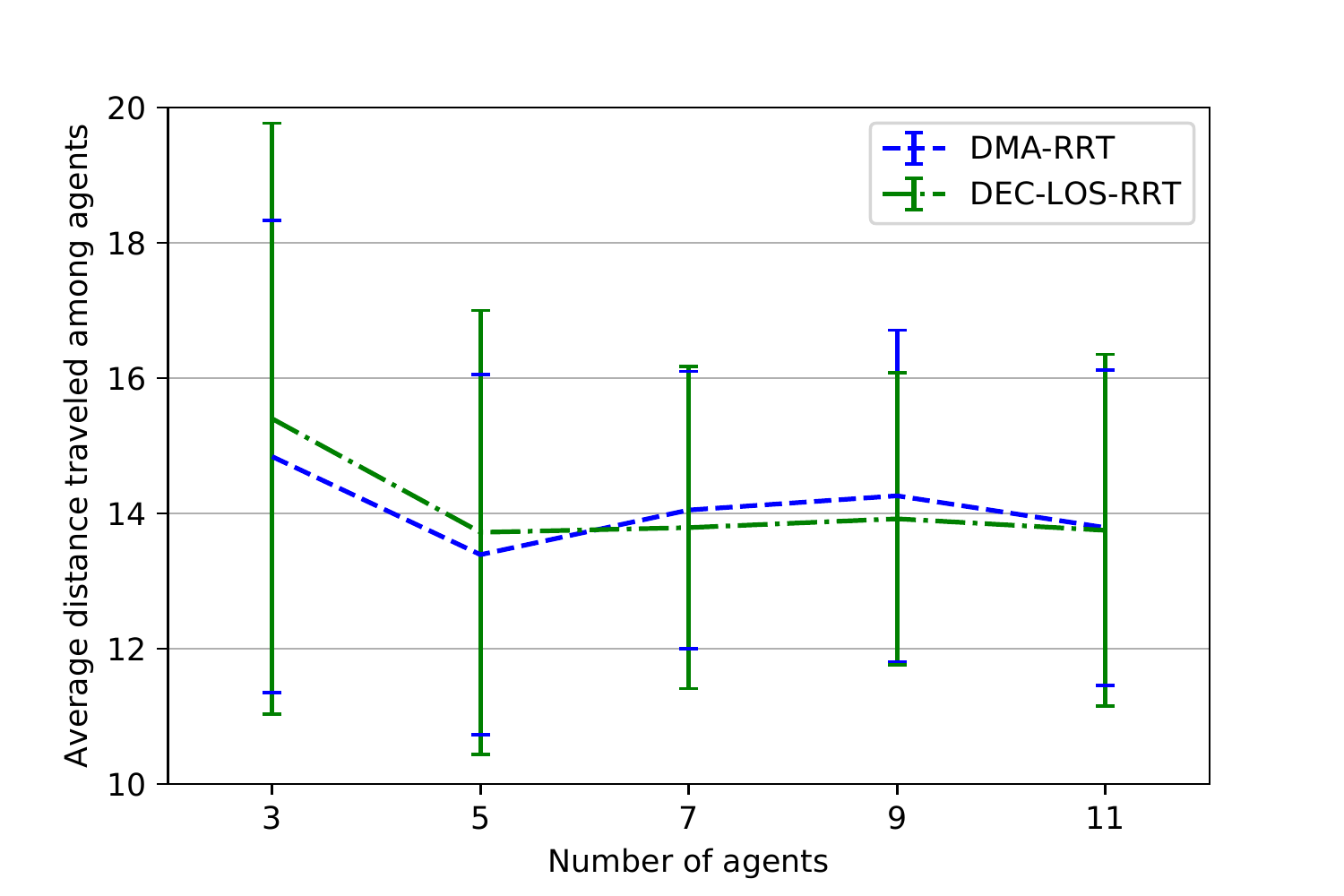}
\vspace{-10pt}
\caption{\small{Average length of final path vs the number of agents running the algorithm. One standard deviation is shown.}}
\label{fig:comparison_length}
\vspace{-5pt}
\end{figure}

For each value of $N=3,5,7$, our approach (DEC-LOS-RRT) results in agents respecting their safety constraints and reaching a goal state within a desired tolerance in this workspace. For $N=9,11$ there were 14 and 17 runs used, respectively, and all but one run for each resulted in all agents finishing. In addition to one setup that resulted in a non-finisher for a DMA-RRT run for $N=9$, these non-finisher setups were excluded from the average calculations in Fig. \ref{fig:comparison_time} and  \ref{fig:comparison_length} for both DMA-RRT and DEC-LOS-RRT. The same setups were used for both algorithms and were randomly generated. Figure \ref{fig:mindists} shows inter-agent distances between every pair of agents for a run with 11 agents. At no point does the distance between agents go below the minimum threshold of $\mindist$.

DMA-RRT and DEC-LOS-RRT are compared in Fig. \ref{fig:comparison_time} and \ref{fig:comparison_length}. Due to the discrete-time implementation of DEC-LOS-RRT, agents cannot stop instantaneously. 0.8m, an increase of 0.2m from 0.6m, was therefore used for $\mindist$ in the implementation of DEC-LOS-RRT because two agents can each travel 0.1m in one time step before they will be able to stop. This addition to $\delta$ can be used in other cases with non-instantaneous stop but constrained velocity and acceleration. We then used the same value of 0.8m for the DMA-RRT implementation. As seen in Figure \ref{fig:comparison_time}, DEC-LOS-RRT when compared with the baseline of DMA-RRT resulted in a longer travel time to goal (when averaged over $N$ agents). This is expected since the baseline DMA-RRT allows for communication between all agents, unlike DEC-LOS-RRT which respects the communication constraints in assumption \ref{assumption:communication}. This increase in time to completion, or time for the agent to travel to the goal state, is due to the agents performing an emergency braking maneuver and recomputing communication subgraphs and paths when another agent that they could not communicate with before now appears in \ac{LOS} as outlined in algorithm \ref{alg:Adjusted_dma_rrt}. It was observed that as the number of agents increases, the average number of emergency braking events goes up. For example, over 16 runs with 3 agents the average number of emergency brake events was 37 per run whereas the average over 16 runs with 11 agents was 155. Given the communication constraints, the increased time to completion is a trade-off our approach makes in order to guarantee safety of all agents at all times.


Figure \ref{fig:comparison_length} shows the average (over $N$ agents) of the path length traveled as the number of agents increase, for both DEC-LOS-RRT and the ideal DMA-RRT baseline.


\subsection{DEC-LOS-RRT with adaptive $\delta$-obstacles}
\label{sec:simresultsadaptive}
\begin{figure}[tb]
\centering
\includegraphics[width=0.49\textwidth, trim={0cm 0cm 0cm 0cm},clip]{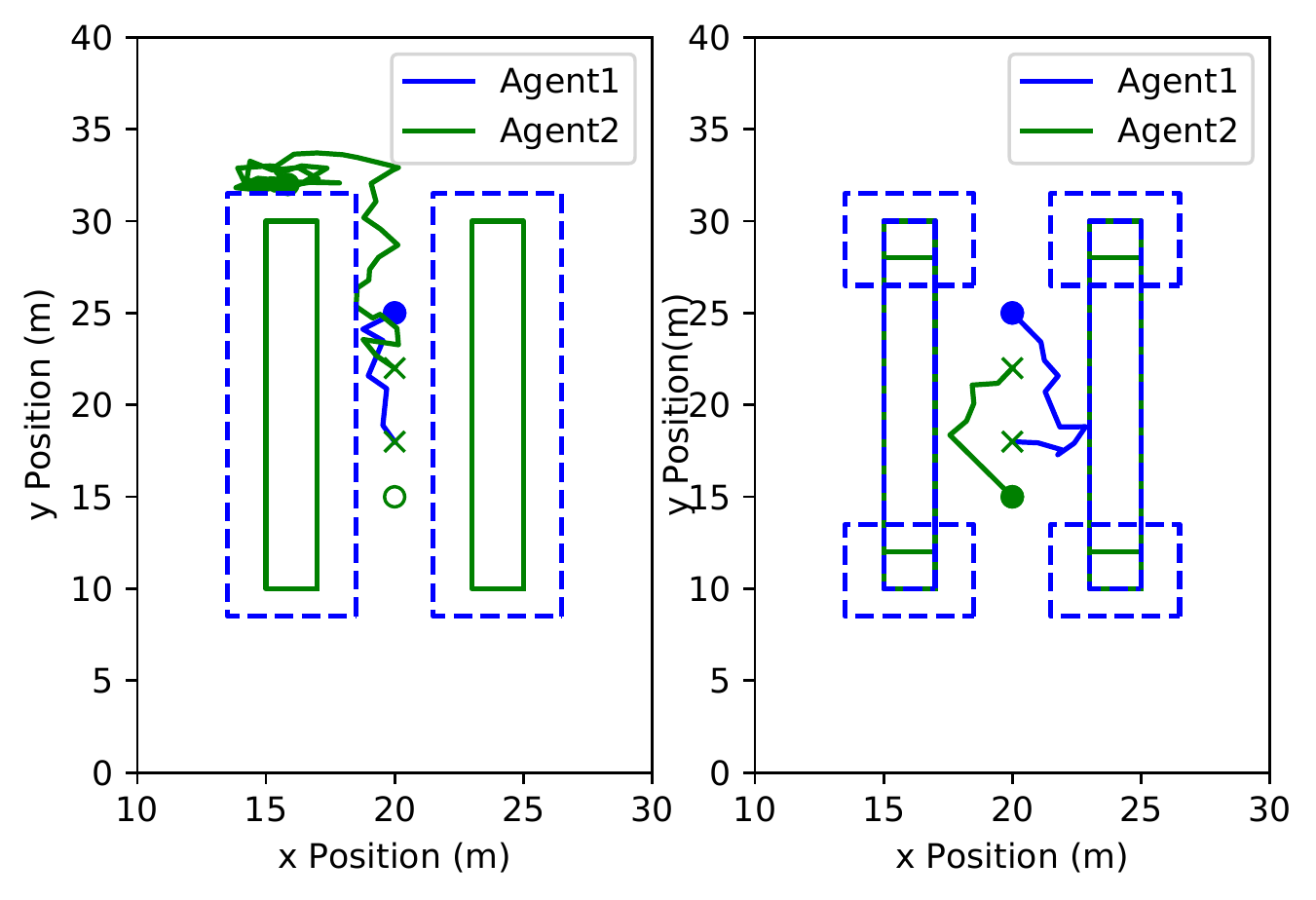}
\vspace{-10pt}
\caption{\small{In this simulation, two agents are trying to move past each other in a corridor. In the left figure, due to the conservative nature of the $\delta$-obstacles, agent 2 is required to move out of the corridor to make progress towards its goal location. In the right figure, the agents plan to avoid the adaptive $\delta$-obstacles, so both agents can instead move past one another in the corridor. \Circle~represents the agent goal states, \CIRCLE~the agent positions at the last time step (coinciding with \Circle~after reaching the goal state), and $\times$ the initial positions.}}
\label{fig:cons_adapt_obstacles_comparison}
\vspace{-5pt}
\end{figure}

As described in section \ref{sec:lessconservative}, it may be more desirable to have a less conservative version of the $\delta$-obstacles. 
One such setting is shown in Fig. \ref{fig:adaptivedelta} and is shown in simulation in Fig. \ref{fig:cons_adapt_obstacles_comparison} where two agents are trying to move past one another in a tight corridor. In these simulations, $\mindist$ is 3m, and start positions were (agent\#-[x,y]): 1-[20, 18], 2-[20, 22] with goal positions 1-[20, 25], 2-[20, 15]. In the more conservative setting with normal $\delta$-obstacles on the left in Fig. \ref{fig:cons_adapt_obstacles_comparison}, agent 2 is forced to move outside of the corridor completely because there is not enough space for both agents to move past each other. However, because agents are memory-less, agent 2 continues to try and circle back to the corridor once it loses sight of agent 1. As it regains sight of agent 1, it realizes it must again plan around the obstacle to reach its goal location, which leads to a \emph{livelock}-like situation where agent 2 repeatedly gets stuck in this process, as is seen by its trajectory on the left in figure \ref{fig:cons_adapt_obstacles_comparison}.

In the adaptive setting on the right in Fig. \ref{fig:cons_adapt_obstacles_comparison}, the agents directly move past each other since there is more room. Both agents still remain safe from any non-visible agent that may move into the corridor because the ends of the obstacles are expanded similar to $\delta$-obstacles (see Lemma \ref{thm:lemma1}).

\subsection{Conclusion}
The simulation results show that our method does indeed guarantee that the safety constraints (section \ref{sec:problemstatement}) are satisfied by each agent at each time step. As compared to the idealized performance of the DMA-RRT baseline, our approach results in the agents requiring a longer time to reach their desired goal positions. As a consequence, as the number of agents increase, DEC-LOS-RRT needs RRT* paths with more nodes otherwise the agents cannot reach their goal state. However, as DEC-LOS-RRT is able to work with a more restrictive communication setup, the trade-off is worthwhile since we still guarantee safety of the agents. We also show that adaptive $\delta$-obstacles may allow for better behavior as they are less conservative with the space. In the future, we will focus on a more rigorous simulation evaluation of this approach, as well as an implementation on actual ground robots.

\acrodef{LOS}{line-of-sight}

\section{Discussion}

\textbf{Summary:} In this work, we have introduced a new algorithm for decentralized multi-agent planning with line-of-sight communication constraints. Our approach builds upon the popular \ac{RRT} as the path planner and guarantees safety of all agents at all times. Simulations show that our approach works well for this multi-agent communication constrained setting. Although we show a slight degradation in performance as compared to a clairvoyant communication setup, we demonstrate how our algorithm provides safety for agents in this more realistic communication setting.

\textbf{Limitations and future work:} The simple dynamics and some of the assumptions (e.g., instantaneous stop) here limit the applicability of this method to some real-world settings, e.g., robots with car-like dynamics. Future work will aim to overcome these limitations, e.g., by using CL-RRT \cite{5175292} as the low-level planner and using the forward reachability of the robot dynamics to plan around obstacles and avoid other robots not in \ac{LOS}. We will also extend our approach to more realistic communication models with latency and inaccuracy. Finally, we will also explore the setting where agents not only transmit their positions to other agents in \ac{LOS} but also a map of their \emph{situational awareness} to avoid the livelock-like situation shown in section \ref{sec:simresultsadaptive}. 

\textbf{Conclusions:} This work presents a first step towards correct-by-construction multi-agent planning with \ac{LOS} communication constraints, and further explorations in this area would focus on adding further realism to both the communication setup as well as robot dynamics.
\section*{ACKNOWLEDGMENT}
This work was supported in part by NSF CPS Frontier project VeHICaL (CNS-1545126). The authors would like to thank Laura Hallock for providing feedback on the paper.


\addtolength{\textheight}{-2cm}

\bibliographystyle{IEEEtran}
\bibliography{refs} 

\appendix
\subsection{Proof sketch for lemma \ref{thm:lemma1}}

\begin{figure}[tb]
\centering
\includegraphics[width=0.49\textwidth, trim={0cm 0cm 0cm 0cm},clip]{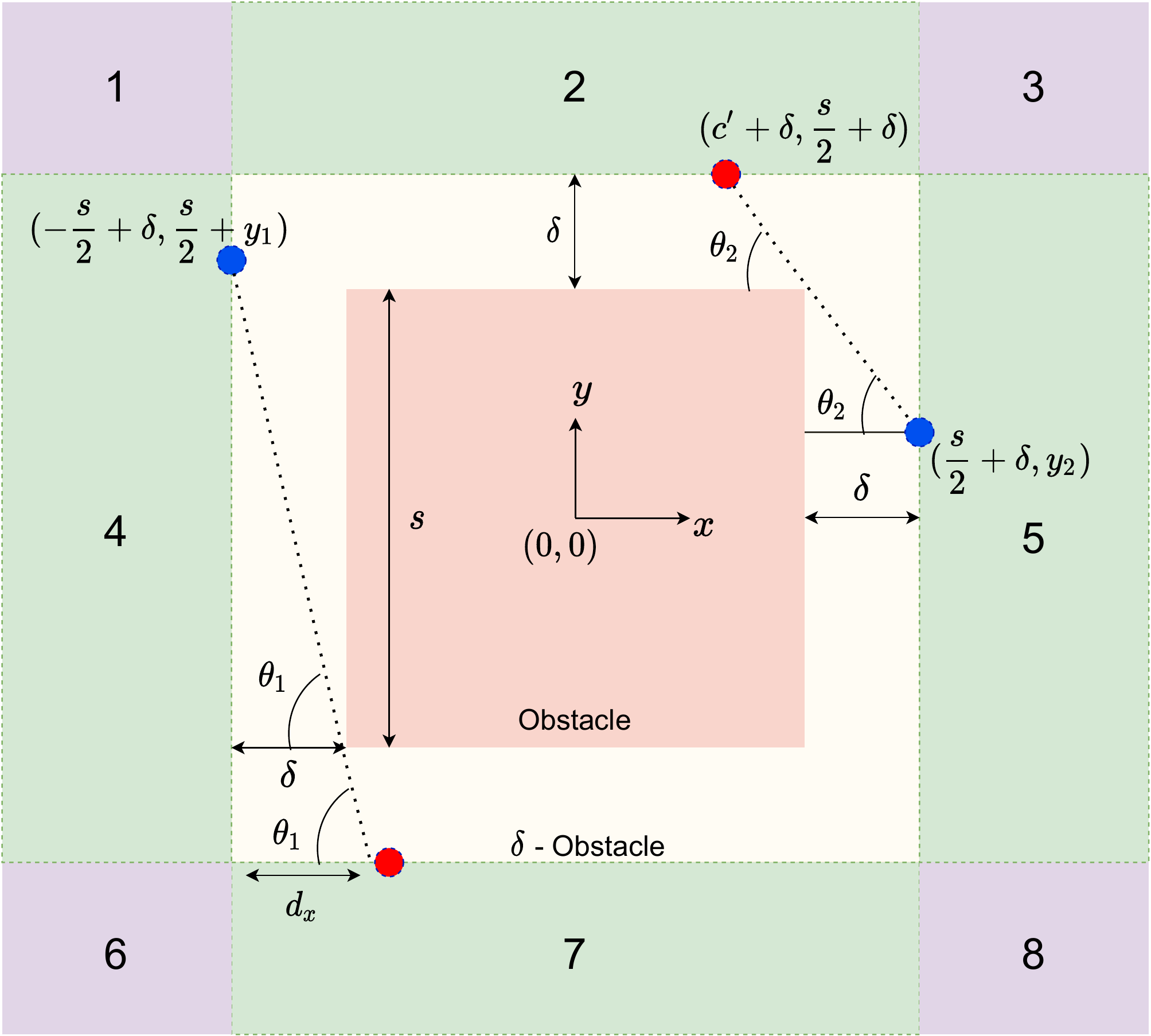}
\vspace{-10pt}
\caption{\small{Visualization of the notable cases in the proof sketch. Without loss of generality, we assume the obstacle is a square.}}
\label{fig:ProofHelper}
\vspace{-5pt}
\end{figure}

The proof is constructed by finding the closest (in the inf-norm sense) that one agent could be to another without being in line-of-sight (LOS), and then proving that this distance is greater than $\mindist$ when we choose a $\delta>(1/2)\mindist$. This is done for multiple cases based on where the two agents are in the workspace, relative to an obstacle. Figure \ref{fig:ProofHelper} shows the obstacle, and the colored regions around it are used to define the cases to be considered. Without loss of generality, and for compactness of notation, we assume the obstacle is a square with side $s$. In all cases shown, the worst-case scenarios involve agents being at the boundaries of the $\delta$-obstacle.

\noindent\textbf{Case 0:} If agent A (in blue in figure \ref{fig:ProofHelper}) is in region 1, the closest that other agent (agent B, show in red) that it is not in line of sight could be is in either regions 5, 7, or 8. For all of these cases, the minimum distance between agent A and B $d_\text{min}^1$ is lower bounded as: $d_\text{min,0} \geq 2\delta+s \geq 2\delta$ 

This is clear from construction, as is shown in the case of agents $A_1$ and $B_1$ shown in figure \ref{fig:ProofHelper}.

\noindent\textbf{Case 1:} Agent A (blue) is in region 4 with a position $(-0.5s-\delta,0.5s+y_1)$ s.t. $y_1 \geq 0$. Agent A has a line of sight to any agent in regions 1, 2, 3, 4 and 6, so the closest agent B (red) that is not in line of sight is in region 7. This case is shown in figure \ref{fig:ProofHelper}. The minimum distance between the y-coordinate of two such agents $d_y \geq \delta+s+y_1$. Let The distance in the x-coordinates of these agents be $d_x = \delta+c$, where $c>0$. We now work out the value of $c$. Let $\theta_1$ be the angle as shown in figure \ref{fig:ProofHelper}, then $\tan \theta_1 = (s+y_1)/\delta = \delta/c$, i.e., $c = \delta^2/(s+y_1)$. The distance between agents A and B, in the inf-norm sense, is $\text{max}[d_x,d_y]=[\delta+\delta^2/(s+y_1), \delta+s+y_1]$. The worst-case minimum of this, over $y_1$, gives us the closest distance between agent A and agent B such that they are not in \ac{LOS} is: $d_\text{min,1} = \text{min}_{y_1 \in (0,\delta]} \text{max}[\delta+\delta^2/(s+y_1), \delta+s+y_1] $.
 Consider the cases $s>\delta-y_1$, $s<\delta-y_1$ and $s=\delta-y_1$. For all three cases, it can be seen that $\text{max}[d_x,d_y] \geq 2\delta$. Hence, $d_\text{min,1} = \text{min}_{y_1 \in [0,\delta]} \text{max}[\delta+\delta^2/(s+y_1), \delta+s+y_1] \geq 2\delta$


\vspace{0.2cm}
\noindent\textbf{Case 2:} Agent A (blue) is in region 5 with a position $(0.5s+\delta, y_2)$, where $y_2 \leq 0.5s$. We again show that the minimum distance between agent A and an agent B (blue) that is not in \ac{LOS} is lower bounded by $2\delta$. For brevity, we work out the case where $y_2\geq0$, but a similar construction applies when $y_2<0$. In this case of interest (see figure \ref{fig:ProofHelper}), the closest that another agent that is not in \ac{LOS} of agent A can be is in region 2. The position of agent B is $(c'+\delta,0.5s+\delta)$, where $c'$ is unknown and will be calculated below.
Let $\theta_2$ be the angle as shown in figure \ref{fig:ProofHelper}, then $\tan \theta_2 = (0.5s-y_2)/\delta = \delta/c'$, i.e., $c' = \delta^2/(0.5s-y_2)$. The minimum distance (in the inf-norm sense), over $y_2$ between agent A and such an agent B in region 2 that is not in \ac{LOS} is therefore $d_\text{min,2}  = \min_{y_2} \text{max}[\delta+\delta^2/(0.5s-y_2), \delta+0.5s-y_2]$. Consider the cases $0.5s-y_2>\delta$, $0.5s-y_2<\delta$ and $0.5s-y_2=\delta$. Again, for all three cases, $d_\text{min,1} \geq 2\delta$.

These cases cover all possible relative positions between an agent A and an agent B that is not in \ac{LOS} of agent A. As seen here, in all cases, the closest these two agents can be while not being in \ac{LOS} is $d_\text{min,i} \geq 2\delta\, \forall i=0,1,2$. Therefore, if $2\delta > \mindist$, then two agents that are not in \ac{LOS} can never come closer than $\mindist$ in the inf-norm sense. This proves the statement of lemma \ref{thm:lemma1}.


\end{document}